\documentclass{article}

\usepackage[preprint]{corl_2026}

\usepackage{booktabs}
\usepackage{amsmath}
\usepackage{amssymb}
\usepackage{tikz}
\usetikzlibrary{arrows.meta,positioning}
\usepackage{graphicx}
\usepackage{tabularx}
\usepackage[export]{adjustbox}

\title{Conceptual Design of an Ecosystem \\
for Real Farm Data Collection toward Agricultural AI Foundation Models}
\author{Junsei Tanaka$^{*1}$, Yoshihiro Sato$^{*1}$\\
$^{*1}$Kyoto University of Advanced Science}
\date{}

\begin{document}
\maketitle

\begin{figure}[h]
\centering
\includegraphics[width=1.05\textwidth, center]{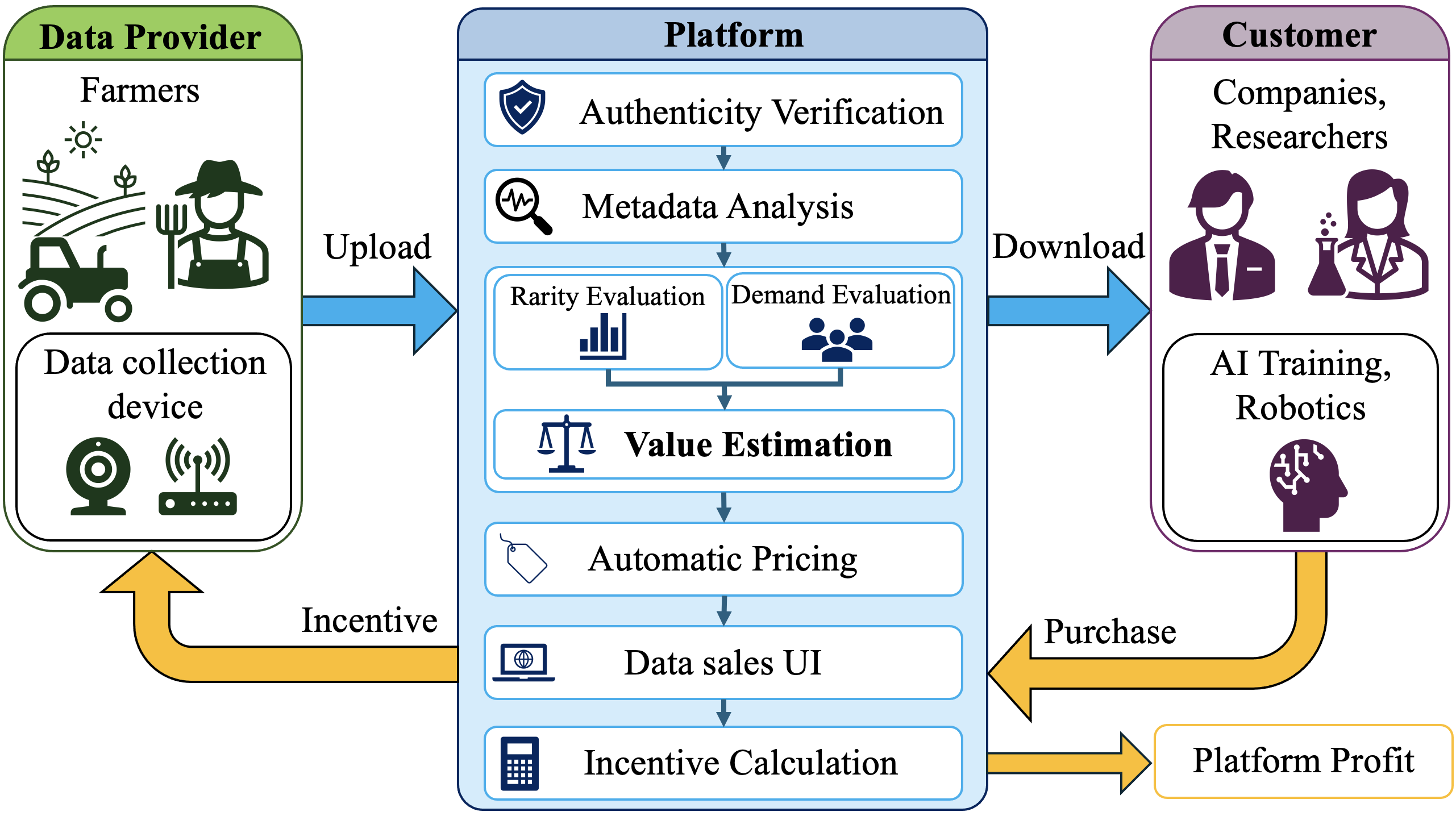}
\caption{Overall ecosystem of the proposed agricultural data collection and distribution platform.
Data providers record video and sensor data during farm operations and upload it to the platform as a sidejob.
The platform verifies data authenticity, performs metadata analysis, and automatically determines a price from rarity and demand evaluation.
Priced data is listed on the marketplace UI for purchase by companies and researchers.
Purchase revenue is shared back to data providers as incentive while the remainder funds platform operations.
This ecosystem makes real farm data collection sustainable by giving farmers an economic reason to keep contributing.}
\label{fig:system_overview}
\end{figure}

\begin{abstract}
Data scarcity is a fundamental challenge in developing AI and foundation models for agricultural robots.
Existing open-source data platforms do not provide sufficient incentives for data providers so long-term data collection remains difficult.
Furthermore, advances in generative AI have introduced a new challenge of verifying that collected data genuinely originates from real farm environments.
We propose an ecosystem for the sustainable collection and distribution of real farm data, integrating automatic pricing driven by demand and rarity, revenue sharing that distributes earnings to farmers as an incentive to keep providing data, and data authenticity guarantees through authenticated device uploads.
To demonstrate the economic sustainability for all three parties among farmers, AI companies, and the platform, we estimate the economic value that agricultural robots stand to generate.
\end{abstract}


\section{Background}

Development of AI and foundation models for agricultural robots requires large volumes of data. However, available public datasets remain limited~\cite{spagnuolo2025agricultural,bommasani2021foundation,heider2025survey}.
Tractor operation recordings capturing weather, road conditions, obstacles, and driver inputs provide rich training signals.
Daily farm footage and environmental sensor data are also important for understanding the agricultural context.
Despite prior efforts, most datasets remain limited to specific crops, farms, seasons, or tasks.
Large-scale data collection targeting agricultural foundation model training is still limited~\cite{heider2025survey,zhu2025harnessing,panda2023agronav,fontani2025fieldrobots}.

Scaling is difficult because on-site data collection in farms is not straightforward.
In-field collection requires coordination with landowners and operators, on-site safety considerations, and constraints imposed by machinery operation~\cite{fontani2025fieldrobots}.
Short-term campaigns cannot capture sufficient diversity because agriculture depends strongly on geography, season, and growth stage~\cite{heider2025survey,zhu2025harnessing}.
Therefore, long-term cooperation from farmers or domain experts is essential~\cite{heider2025survey,panda2023agronav}.

To maintain long-term farmer participation, a mechanism that returns fair value for their contributions is necessary~\cite{wiseman2019farmers,vanderburg2021trust,jouanjean2020oecd,sullivan2024trust}.
Such a mechanism requires not only data storage but also a distribution infrastructure including pricing, transactions, revenue sharing, and trust assurance.
Agricultural data value varies widely with content.
Automated pricing is therefore important, yet no established mechanism for pricing data based on its content exists in agriculture~\cite{zhang2024datamarkets,zhang2023pricing}.
Building such a platform also requires addressing data authenticity.
Advances in generative AI have made it increasingly critical to verify that collected footage genuinely originates from farm environments~\cite{openai2024sora,bommasani2021foundation}.

Prior work has studied dataset preparation, data sharing rules and management, and authenticity assurance as separate problems~\cite{heider2025survey,fontani2025fieldrobots,panda2023agronav,wiseman2019farmers,vanderburg2021trust,jouanjean2020oecd,sullivan2024trust,manoj2023trusted}.
No prior work integrates continuous farmer compensation, content-based dynamic pricing, and data authenticity guarantees for real farm data collection in agricultural robotics~\cite{zhang2024datamarkets,zhang2023pricing}.

\section{Purpose}

This paper proposes a conceptual design of an ecosystem for the sustainable collection and distribution of real farm data for training agricultural AI foundation models.
The ecosystem centers on an automatic pricing algorithm reflecting demand and rarity. It is further integrated with mechanisms for continuous farmer revenue and data authenticity assurance.

Success is defined by two criteria.
First, the design must simultaneously satisfy automatic pricing reflecting demand and rarity, continuous farmer revenue through data transactions, and data authenticity guarantees.
Second, we estimate the economic value generated by agricultural robots to demonstrate economic sustainability for all three parties.

The scope covers data collected through agricultural activities. This includes camera, environmental sensor, and GPS data from agricultural machinery as well as farm work recordings.
In the proposed ecosystem, farmers serve as data providers and agricultural AI researchers and companies serve as data buyers.
This paper presents a conceptual design. Therefore, robustness validation, large-scale simulation, and implementation are left as future work.

\section{Related Works}

This section reviews related work and clarifies the positioning of this study.
We begin with the most closely related examples of agricultural data markets, then organize broader related work in Table~\ref{tab:related_work_positioning}.

\textbf{Farmobile}~\cite{farmobile2021ownership} is an American agricultural technology company that helps farmers collect and sell their field data.
It connects sensors attached to agricultural machinery to collect operational data such as speed, fuel consumption, and seeding rate in the cloud.
Farmers can list their recorded data on a data store and earn revenue at a fixed per-acre rate with a 50\% farmer share.
In contrast to our automatic pricing based on demand and rarity with equal revenue distribution, Farmobile uses a fixed-price, individual-contract business model, and the incentive structure and pricing mechanism differ fundamentally.

\textbf{WAGRI}~\cite{wagri2019} is a public agricultural data infrastructure operated by the National Agriculture and Food Research Organization in Japan.
It aggregates data via API from multiple providers including agricultural machinery manufacturers, weather services, the Ministry of Agriculture, Forestry and Fisheries~\cite{maff_website}, and agricultural cooperatives.
However, WAGRI provides no mechanism for farmers to list or sell their own recorded data and includes no incentive or pricing functions.

Additional related work is organized in Table~\ref{tab:related_work_positioning}.

\begin{table*}[t]
\centering
\caption{Positioning of related work}
\label{tab:related_work_positioning}
\renewcommand{\arraystretch}{1.2}
\begin{tabularx}{\linewidth}{p{2.3cm} p{3.2cm} p{2.0cm} X}
\hline
Category & Primary objective & Representative references & Relation to this work \\
\hline
Agricultural data markets \& sharing examples
& Building mechanisms for agricultural data collection, sharing, and sale
& \cite{farmobile2021ownership,phillips2019digital,borrero2022platform,baarbe2019commons}
& Most closely related, but not designed around real farm data collection for agricultural robots with rarity-based pricing and authenticity guarantees integrated. \\
\hline
Agricultural data sharing \& governance
& Clarifying trust and rules for agricultural data sharing
& \cite{wiseman2019farmers,vanderburg2021trust,ryan2024policy,duncan2026political,sullivan2024trust,atik2022ownership,chamorro2025sharing}
& Important for trust, rights, contracts, and policy, but does not primarily address concrete pricing or revenue-sharing mechanisms. \\
\hline
General data markets \& pricing
& Designing pricing, distribution, and incentives for data transactions
& \cite{zhang2024datamarkets,zhang2023pricing,gzhang2025pricingreview,azcoitia2022measuring,jiang2025incentive,ezhang2025modelsharing}
& Theoretically relevant for pricing and distribution design, but not based on field constraints of agriculture or sustained farmer cooperation. \\
\hline
Authenticity assurance \& data origin
& Guaranteeing the origin and integrity of data
& \cite{manoj2023trusted,bergier2025agritrust,bergier2026framework,jang2025authentication,hasan2024assurance,guo2025bigdata}
& Related in terms of data trustworthiness, but does not integrate market design, pricing, and revenue sharing. \\
\hline
Agricultural datasets \& robotics
& Building data and technology foundations for agricultural AI and robots
& \cite{heider2025survey,spagnuolo2025agricultural,fontani2025fieldrobots,hughes2015plantvillage,chiu2020agrivision,panda2023agronav,jeong2025agrichrono}
& Important foundations, but do not treat continuous data collection, farmer compensation, or distribution infrastructure as primary topics. \\
\hline
External comparative references
& Illustrating contrast with large-scale or commercial data collection mechanisms
& \cite{sorokin2008mturk,vanhorn2018inaturalist,caesar2020nuscenes,ettinger2021waymo,scaleai2024}
& Useful references for large-scale and commercial data collection, but not based on difficult-access farm environments or farmer revenue sharing. \\
\hline
\end{tabularx}
\end{table*}

Research integrating agricultural data pricing with farmer revenue sharing in the context of data collection for agricultural robots remains limited~\cite{wiseman2019farmers,vanderburg2021trust,jouanjean2020oecd,sullivan2024trust,zhang2024datamarkets,manoj2023trusted}.
This work proposes a conceptual design unifying pricing, revenue distribution, and authenticity assurance.

\section{Proposed Ecosystem Concept and Design}

This section describes the proposed ecosystem.
Section~4.1 presents the overall structure and self-reinforcing properties.
Section~4.2 describes the automatic pricing algorithm.
Section~4.3 covers the incentive design.
Section~4.4 details the data authenticity guarantee.
These components create a structure in which farmers earn continuous revenue simply by continuing their agricultural work. As a result, data provision becomes a natural side income.

\subsection{Ecosystem Overview}

The ecosystem has two types of participants. Farmers serve as data providers while AI researchers and companies serve as data buyers. The platform serves as the intermediary (Figure~\ref{fig:system_overview}).

Farmers attach the platform-provided fully automated data collection device to their machinery.
The device records footage automatically during farm operations and uploads it to the platform.
Upon upload, a digital signature verifies data authenticity.
Authenticated data is enriched with metadata on detected objects, events, and conditions using a vision-language model~\cite{qwen3vl2025}.
Rarity scores are computed from this metadata and fed into the pricing algorithm.
Buyers browse metadata and sample previews and purchase relevant recordings.
Purchase counts feed back into demand scores and the incentive calculation. Residual revenue funds platform operations.

\subsubsection{Self-Reinforcing Cycle}

\begin{figure}[h]
\centering
\includegraphics[width=0.95\linewidth]{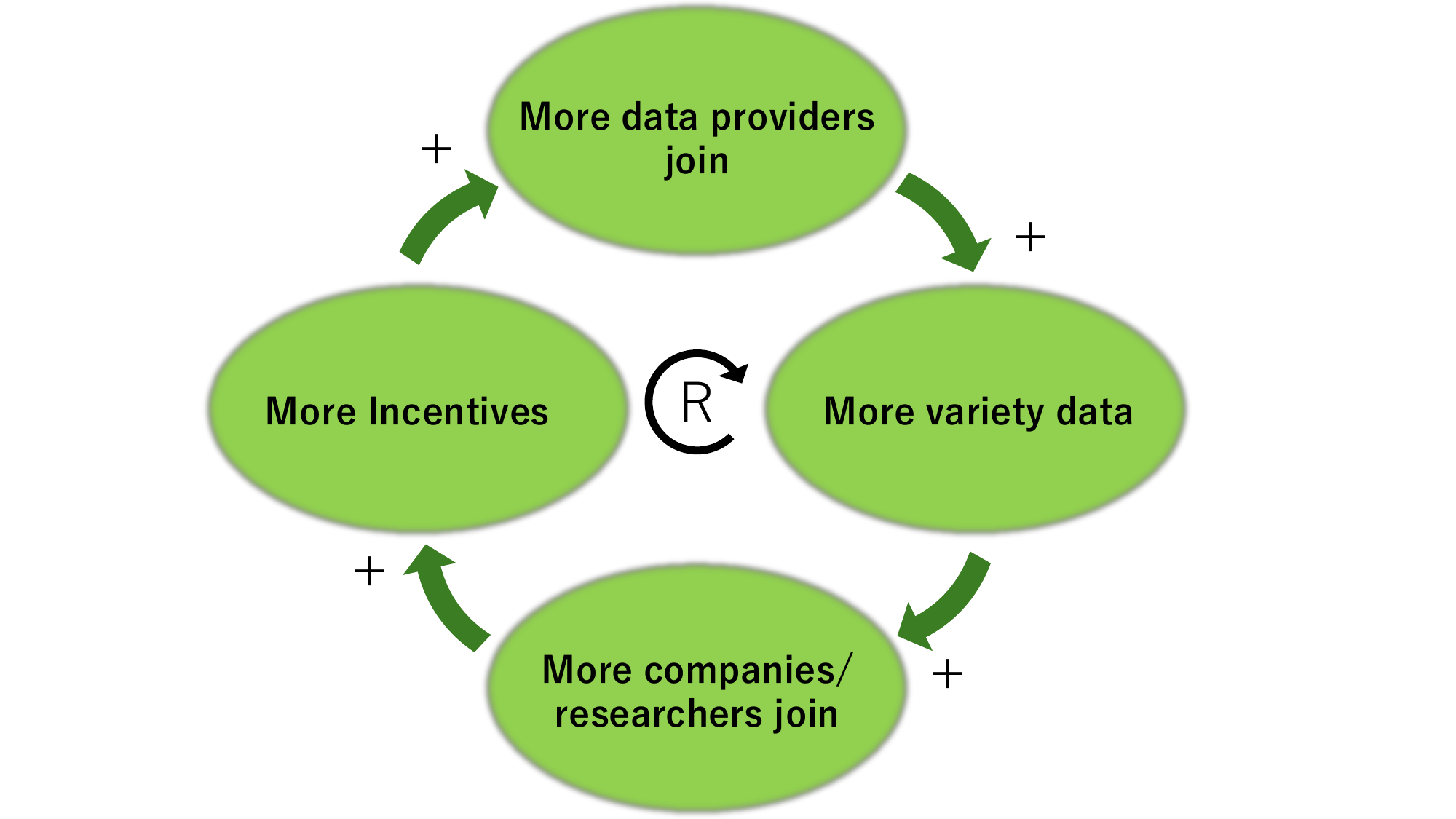}
\caption{Self-reinforcing cycle of the proposed ecosystem.
         Farmer participation, data diversity, buyer demand, and farmer revenue each strengthen the next to drive sustained ecosystem growth.}
\label{fig:causal_loop}
\end{figure}

The revenue-sharing mechanism incentivizes data provision. This self-reinforcing cycle is illustrated in Figure~\ref{fig:causal_loop}.
Data diversity grows as more farmers join. This attracts buyers with varied research needs.
Greater data volume improves rarity score accuracy and refines valuation.
Higher purchase volume increases farmer revenue and draws further participants.
This cycle gives the ecosystem self-reinforcing properties, and sustainability emerges naturally from its structure.

\subsection{Automatic Pricing Algorithm}

The following describes one possible implementation of the automatic pricing algorithm.
Each recorded session is priced automatically based on demand and rarity.
$P_{\min}$ is a fixed floor price set by the platform. A ceiling price can also be configured.
The selling price $P$ is:
\begin{equation}
  P = P_{\min} \times (1 + d + R)
  \label{eq:price}
\end{equation}
The constant term $1$ in $(1 + d + R)$ ensures the price never falls below $P_{\min}$ when $d = 0$ and $R = 0$.
Demand and rarity are reflected as independent additive terms.
The price does not depend on recording duration. A long recording with no purchases and no rare events stays at the floor price, while a short recording containing many rare events commands a high price.

\subsubsection{Demand Evaluation}

The demand counter $d$ is an integer tracking recent purchase activity.
It increments by 1 when a purchase occurs within the platform-configured lifetime $T_{\text{life}}$. It decrements by 1 when no purchase occurs beyond $T_{\text{life}}$ and never falls below zero.
Continuously purchased recordings accumulate higher $d$. Demand cools gradually as interest wanes.
In practice, $d$ is expected to remain on a similar scale as typical $R$ values (roughly 0--10).

\subsubsection{Rarity Evaluation}

A VLM analyzes the metadata of uploaded footage and detects specific events~\cite{min2025cv_generative_agriculture}.
An event is a notable occurrence within a specific time segment of the footage. Each event type is registered as a category on the platform.
For example, a recording that contains 30 seconds of animal intrusion adds those 30 seconds to the cumulative duration $\tau(e)$ of that event type.
The rarity score $R$ is then computed from these event durations.
The occurrence probability of event $e$ is defined as its share of total recording time across all platform recordings:
\begin{equation}
  p(e) = \frac{\tau(e)}{T_{\text{total}}}
\end{equation}
where $\tau(e)$ is the total duration of event $e$ across all recordings and $T_{\text{total}}$ is total recording duration on the platform.
The rarity score is:
\begin{equation}
  R = \sum_{j} k_j \cdot \bigl(-\log\, p(e_j)\bigr)
\end{equation}
where $j$ indexes all event types detected in the recording and $k_j$ is the number of scenes where event type $e_j$ occurs.
For example, when $p(e) = 100\%$, $1\%$, $0.001\%$, $R \approx 0$, $4.6$, $11.5$ respectively.
With $P_{\min} = 1{,}000$ JPY and $d = 0$, prices become 1,000, 5,600, and 12,500 JPY.

\subsection{Incentive Design}

Farmers receive incentives through revenue sharing.
The distribution amount $S_u$ for each uploader $u$ is:
\begin{equation}
  S_u = \frac{\rho \cdot M_{\text{total}}}{U_{\text{active}}}
  \label{eq:incentive}
\end{equation}
where $M_{\text{total}}$ is total platform revenue, $\rho$ with $0 < \rho \leq 1$ is the fraction distributed to all farmers, and $U_{\text{active}}$ is the number of uploaders who submitted at least one recording during the distribution period.
The Platform Profit $(1-\rho) \cdot M_{\text{total}}$ is used for platform maintenance and operations.
Participants who do not upload anything during a period are excluded from distribution to prevent free-riding.

\subsection{Device Authentication and Upload Verification}

The ecosystem uses device authentication to prevent unauthorized uploads and data tampering.

\textbf{Device authentication:}
First, a device-specific key pair is generated during manufacturing and the private key is stored in a Secure Element~\cite{maene2018trusted} protected against tampering and readout.
Next, the platform adds the public key to a whitelist of devices when a farmer registers the data collection device.
At upload, the data collection device signs the data with its private key~\cite{nist_fips186_5} and sends it to the platform. The platform then verifies the signature against the registered public key and rejects uploads from unknown devices or tampered data.

\textbf{Buyer fraud reports:}
A VLM reviews each upload and rejects footage that does not show a real farm environment.
In addition, buyers can report suspicious uploads to the platform. The uploader's account is suspended if the platform confirms misconduct.

\section{Economic Sustainability and Pricing Dynamics}

\subsection{Economic Returns for All Three Parties}

This chapter demonstrates that farmers, AI companies, and the platform can all achieve positive returns.
Farmers earn revenue when data is purchased. The platform retains revenue after deducting operating costs.
Since AI company sustainability depends on robot business scale, we estimate the economic value generated by agricultural robots and show that data purchase costs can be covered.

Agricultural robots that replace farm labor would eliminate labor costs.
Additionally, continuous monitoring and precision agriculture enabled by AI and robotics would generate additional value through yield improvement, reduced chemical fertilizer costs, reduced weather disaster losses, reduced pest damage, and reduced wildlife damage.
These economic values are estimated for Japan, where detailed agricultural statistics are available. Results are shown in Table~\ref{tab:market_potential}.

\begin{table*}[t]
\centering
\caption{Potential economic value of agricultural robots and the data ecosystem (Japan)}
\label{tab:market_potential}
\renewcommand{\arraystretch}{1.2}
\begin{tabular}{p{4.5cm} r p{7.5cm}}
\toprule
Item & Max.\ potential & Basis \\
\midrule
\textbf{Labor substitution} & 2.2051T JPY
  & (Agricultural income + hired labor + work clothing) avg.\ 2,543K JPY/farm $\times$ 867,045 operations~\cite{maff_einou_r6,maff_census_2020} \\[4pt]
\textbf{Yield improvement} & 514.4B JPY
  & Crop output (5.7266T JPY) $\times$ 9\%~\cite{maff_seisan_r5,maff_smartagri_project} \\[4pt]
\textbf{Fertilizer cost reduction} & 148.8B JPY
  & Total fertilizer costs ($\sim$500B JPY) $\times$ 30\%~\cite{maff_smartagri_project} \\[4pt]
\textbf{Weather disaster reduction} & 243.7B JPY
  & Average of MAFF Annual Report figures for FY2020~\cite{maff_disaster_r2} and FY2022~\cite{maff_disaster_r4} \\[4pt]
\textbf{Pest damage reduction} & 1.1453T JPY
  & Crop output $\times$ FAO~\cite{fao_pest_damage} lower bound 20\% \\[4pt]
\textbf{Wildlife damage reduction} & 18.8B JPY
  & MAFF Wildlife Crop Damage Statistics, FY2024 national total~\cite{maff_wildlife_damage} \\[4pt]
\midrule
\textbf{Total} & \textbf{$\sim$4.3T JPY} & \\
\bottomrule
\end{tabular}
\end{table*}

All figures represent maximum estimates. Even partial realization is sufficient for three-party sustainability.
Although this analysis targets Japan, the platform is applicable worldwide.
Japan's agricultural output accounts for approximately 2.7\% of global output~\cite{fao_world_agri}, suggesting global market potential could be hundreds of times larger.

\textbf{Three-party sustainability:}

\begin{itemize}
  \item \textbf{Farmers}: Farmers currently provide data at no cost. The platform gives them income by continuing their farm work.
  \item \textbf{AI companies}: Table~\ref{tab:market_potential} represents the maximum economic value from agricultural automation by robots and does not directly indicate data market size, but suggests that data purchase costs can be covered.
  If robot companies capture even 1\% of this value as service revenue, roughly 43B JPY, data purchase costs are far smaller. Sustainability is established.
  Data costs exceeding total service revenue is not a realistic scenario.
  \item \textbf{Platform}: Data sales revenue funds platform operations. Target agricultural operations number approximately 870,000 entities in Japan. Sustained operations at this scale are feasible.
\end{itemize}

\subsection{Pricing Algorithm Dynamics}

\begin{figure}[h]
\centering
\includegraphics[width=0.48\linewidth]{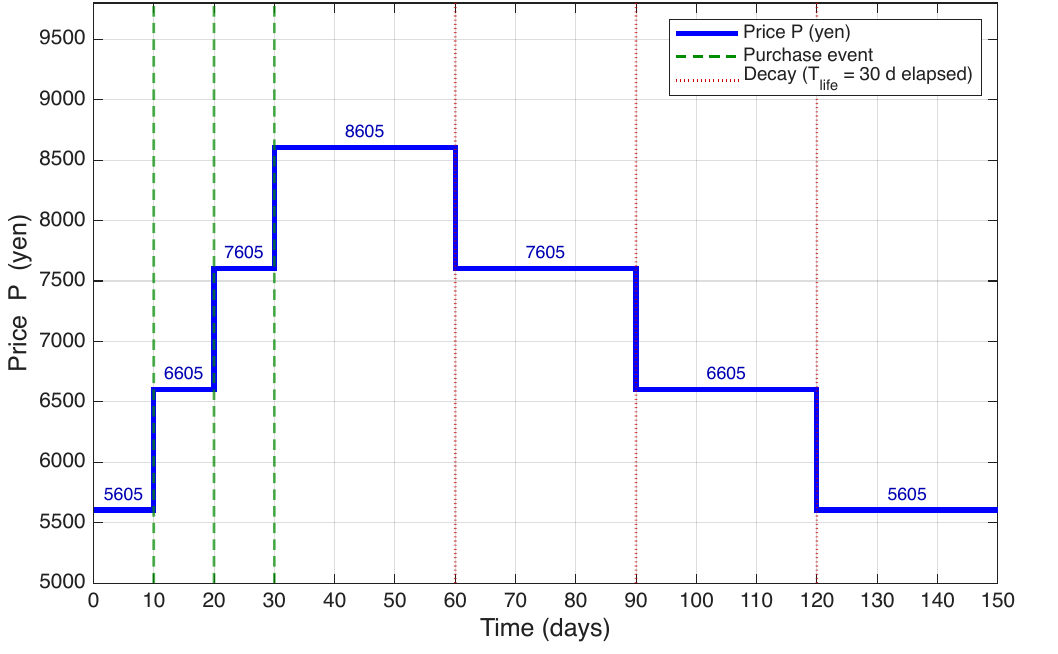}
\caption{Price dynamics under $T_{\text{life}} = 30$ days, $P_{\min} = 1{,}000$ JPY, $p(e) = 1\%$, with purchases on Days 10, 20, and~30.
         The price rises with each purchase and converges stepwise back to the base price over successive inactive periods.}
\label{fig:price_dynamics}
\end{figure}

To illustrate how prices evolve in practice, we present a concrete example.
For data containing a rare event with $p(e) = 1\%$, when purchases occur on Days~10, 20, and~30, the demand counter $d$ rises step by step and then gradually returns to the base price after each $T_{\text{life}} = 30$-day inactive period, as shown in Figure~\ref{fig:price_dynamics}.

\section{Discussion}

\textbf{Corporate Exploitation of Agricultural Data:}
AI companies are attempting to collect agricultural data for free and build machines to replace farm workers.
This is obviously exploitation and ethically wrong.
Agricultural data has no market price and can be collected without farmers knowing.
Farmers have no way to know that their data will one day be used to replace their labor.
Moreover, companies do not tell them.
In this way, farmers tighten the noose around their own necks with their own data without knowing it.
The costs of this automation fall on all of society while profits go only to corporations.
Ultimately, if corporations control food production, all of humanity will have no choice but to follow their terms.
This is a threat to human freedom itself.

This is why farmers must act now to secure rights over their own data.
As long as data has a price, farmers have the power to choose which companies to sell to or to refuse.
Whether farmers become victims or active parties who shape their own future is decided now at this stage of data collection.

This platform is one approach to solve this problem.
It aims to prevent corporate data monopoly and return control to farmers by pricing agricultural data and giving farmers revenue.
This ecosystem is built so farmers can negotiate before automation advances.

\textbf{Expansion of Demand-Supply Matching:}
In the current ecosystem, farmers automatically record data through their agricultural activities.
Adding a request function through which data buyers can directly present to farmers the types, conditions, and target prices of data they need would enable bidirectional demand-supply matching.
Farmers would gain more concrete understanding of the value of the data they record, and buyers could more reliably acquire the data they need.
Such an extension would further enhance the usefulness of the ecosystem.

\textbf{Defensive Protection of Farm Landscape:}
Participation in this platform may also give farmers a means of defending their farm landscape against unauthorized use.
By selling farm footage as a commercial data product, farmers create a documented record that their landscape has recognized commercial value.
This record may support claims against unauthorized recording of their farm or the use of such footage as AI training data without consent.
The platform thus offers not only income but also a means of protecting the farming environment itself.

\section{Summary}

Long-term farmer cooperation is essential for collecting real farm data to train agricultural AI foundation models, yet no established mechanism supports this.
Automated data pricing and authenticity verification are also unresolved challenges.
This paper proposes an ecosystem integrating automatic pricing, incentive design, and data authenticity guarantees.
Dynamic pricing based on rarity and demand, a revenue-sharing model, and digital-signature-based authenticity assurance are described as core components.
Economic sustainability for all three parties is demonstrated through an estimate of the market potential of agricultural automation by robots in Japan.

Farmers can earn extra income through their ordinary agricultural activities, and AI researchers and companies can obtain diverse real farm data at transparent prices.
The fundamental barrier to real farm data collection lies in the absence of an appropriate ecosystem. Designing it so farmers are among those who benefit is key to sustainable data collection.
This is simultaneously a technical challenge and an ethical responsibility.

\textbf{Future work.}
Implementation and simulation to quantitatively validate the proposed ecosystem are the highest priority. The convergence of the pricing algorithm, rarity score accuracy, and the effect of incentives on farmer participation must be evaluated.
The applicability of these design principles to other difficult-access environments such as forestry, fisheries, and construction sites is also left as future work.


\end{document}